\documentclass[10pt,twocolumn,letterpaper]{article}

\usepackage{3dv}
\usepackage{times}
\usepackage{epsfig}
\usepackage{graphicx}
\usepackage{amsmath}
\usepackage{amssymb}

\usepackage{algorithm}
\usepackage{algorithmicx}
\usepackage{algpseudocode}

\usepackage{booktabs}
\usepackage{tabularx}
\newcolumntype{Y}{>{\centering\arraybackslash}X}
\usepackage{xspace}
\usepackage{appendix}

\usepackage[pagebackref=true,breaklinks=true,letterpaper=true,colorlinks,bookmarks=false]{hyperref}

\threedvfinalcopy

\ifthreedvfinal\fi
\begin{document}

\title{Global Context Aware Convolutions for 3D Point Cloud Understanding}

\newcounter{auxFootnote}
\renewcommand{\thefootnote}{\fnsymbol{footnote}}

\author{Zhiyuan Zhang$^1$ \quad\quad Binh-Son Hua$^{2,3,\footnotemark\setcounter{auxFootnote}{\value{footnote}}}$ \quad\quad Wei Chen$^1$ \quad\quad Yibin Tian$^{1,\footnotemark[\value{auxFootnote}]}$ \quad\quad Sai-Kit Yeung$^{4,\footnotemark[\value{auxFootnote}]}$
	\vspace{0.2cm}\\
	$^1$Litemaze \quad\quad\quad $^2$VinAI Research, Vietnam \quad\quad\quad $^3$VinUniversity, Vietnam\\
	$^4$Hong Kong University of Science and Technology
	\vspace{-0.2cm}\\
}

\maketitle

\def\ournet{GCANet\xspace}
\def\ourconv{GCAConv\xspace}
\def\smallgap{\vspace{0.05in}}
\begin{abstract}
Recent advances in deep learning for 3D point clouds have shown great promises in scene understanding tasks thanks to the introduction of convolution operators to consume 3D point clouds directly in a neural network. 
Point cloud data, however, could have arbitrary rotations, especially those acquired from 3D scanning. Recent works show that it is possible to design point cloud convolutions with rotation invariance property, but such methods generally do not perform as well as translation-invariant only convolution. 
We found that a key reason is that compared to point coordinates, rotation-invariant features consumed by point cloud convolution are not as distinctive. 
To address this problem, we propose a novel convolution operator that enhances feature distinction by integrating global context information from the input point cloud to the convolution. 
To this end, a globally weighted local reference frame is constructed in each point neighborhood in which the local point set is decomposed into bins. Anchor points are generated in each bin to represent global shape features. A convolution can then be performed to transform the points and anchor features into final rotation-invariant features. 
We conduct several experiments on point cloud classification, part segmentation, shape retrieval, and normals estimation to evaluate our convolution, which achieves state-of-the-art accuracy under challenging rotations.
\end{abstract}

\footnotetext[\value{auxFootnote}]{Corresponding author}

\section{Introduction}
\label{indroduction}
Scene understanding has long been a challenging problem in computer vision. Recently, there have been significant advances in applying deep learning~\cite{lecun2015deep} to train neural networks for numerous tasks such as object classification and semantic segmentation. With the wide availability of consumer-grade depth sensors, acquiring 3D data has become more intuitive and robust with many 3D datasets available publicly~\cite{wu-3dshapenets-cvpr15,chang2015shapenet,hua2016scenenn,dai2017scannet,armeni-parsing-cvpr16,yi2016scalable,uy-scanobjectnn-iccv19}. This leads to increased interests in tackling scene understanding in the 3D domain. 

Among the representations for 3D data, a promising direction is to let neural networks consume point cloud data directly since point cloud data is the common data format acquired from depth sensors such as RGB-D or LiDAR cameras. However, since a point cloud is a mathematical set and so it fundamentally differs from an image, passing a point cloud to a traditional neural network like those in the image domain does not work. In principle, it is necessary to design a convolution-equivalent operator in the 3D domain that can take a point cloud as input and output its per-point features. Several attempts have been made with promising results~\cite{qi2017pointnet,qi2017pointnet++,hua2017point,li2018pointcnn,xu2018spidercnn,zhang-shellnet-iccv19}. 

Despite such research efforts, a problem often overlooked in point cloud convolution is that the operator does not exhibit rotation invariance. A viable solution in 2D deep learning is to augment training data with random rotations. However, in 3D, such data augmentation becomes less effective due to the additional degree of freedom in representing 3D rotations, which can make training prohibitively expensive. A few works turn to learn rotation-invariant features~\cite{zhang-riconv-3dv19,rao-spherical-cvpr19,poulenard-spherical-3dv19,deng2018ppf,chen2019clusternet}, which allows consistent predictions given arbitrarily rotated point clouds. 

Unfortunately, a limitation from previous works is that rotation-invariant convolution does not yield features that are as distinctive as translation-invariant convolution. This makes performing object classification with aligned data more accurate than performing the same task with data with arbitrary rotations. For exact rotation invariance, it is expected that the rotation-invariant convolution is as accurate as its translation-invariant sibling. 

In this paper, we propose a novel approach for performing rotation-invariant convolution for point clouds. Our key observation is that when rotation invariance is added, it introduces some ambiguities and thus reduces feature distinctiveness. To address this problem, we propose to integrate global context information from the input point cloud to the convolution, resulting in a global context aware convolution for 3D point clouds. 
The main contributions of this work are:
\begin{itemize}
\item \ourconv, a novel rotation-invariant convolution operator that output features from local point sets and global anchors. Each anchor is built from subdivided spaces using a globally-weighted local reference frame at each keypoint. By explicit encoding the relation between local point sets and the global anchors, \ourconv can capture both local and global context; \vspace{-0.1in}

\item \ournet, a neural network architecture that uses \ourconv for learning rotation-invariant features for 3D point clouds. The network allows consistent performance across training/testing scenarios that involves different rotation modes; \vspace{-0.1in}

\item Applications of \ournet on object classification, object part segmentation, shape retrieval, and normals estimation that achieve the state-of-the-art performance under challenging rotations.
\end{itemize}
\section{Related Works}
\label{related_works}

Deep learning in the 2D domain has witnessed great success in solving scene understanding tasks such as object classification, semantic segmentation, normal estimation, etc. Drawing from this inspiration, techniques for deep learning in the 3D domain has recently been developed with promising results. In this section, we review the state-of-the-art research in deep learning with 3D data, and then focus on techniques that enable feature learning on point clouds for scene understanding tasks. 

Early research in 3D deep learning focus on regular and structured representations of 3D scenes such as multiple 2D images~\cite{su2015multi,qi2016volumetric,esteves2019equivariant}, 3D volumes \cite{qi2016volumetric,li2016fpnn}, hierarchical data structures like octree \cite{riegler2017octnet} or kd-trees \cite{klokov2017escape,wang2017cnn}. Such representations yield good performance. However, they face challenges from a practical point of view due to memory consumption, imprecise representation, or lack of scalability when high-resolution data is employed. 

Many recent works in 3D deep learning switched to investigate how to learn with 3D point cloud, a more compact and intuitive representation compared to volumes and image sets. However, performing deep learning with 3D point clouds is not as straightforward as extending 2D image convolution to 3D because mathematically, a point cloud is a set. To define a valid convolution for a point cloud, it is necessary to ensure that the output features from a convolution is invariant to the permutation of the point set. PointNet~\cite{qi2017pointnet} pioneered such a solution to output global features by maxpooling per-point features from MLPs. Several follow-up works focus on designing convolutions that can learn local features for a point cloud efficiently~\cite{hua2017point,qi2017pointnet++,li2018pointcnn,xu2018spidercnn,wang2018edgeconv,zhang-shellnet-iccv19}. Please also refer to the technical report by Guo et al.~\cite{guo-point-survey-2019} for further summary of many deep learning techniques for 3D point clouds. 

A fundamental missing feature in the previously mentioned convolution for point clouds is that rotation invariance is not supported. A common solution is to augment the training data with arbitrary rotations, but a limitation of doing so is that generalizing the predictions to unseen rotations is challenging, not mentioning that the training time becomes longer due to the increased amount of training data. Instead, it is desirable to have a point cloud convolution with rotation-invariant features. 

To this end, Rao et al.~\cite{rao-spherical-cvpr19} map a point cloud to a spherical domain to define a rotation-invariant convolution.
Zhang et al.~\cite{zhang-riconv-3dv19} proposed a convolution that operates on features built from Euclidean distances and angles. Poulenard et al.~\cite{poulenard-spherical-3dv19} proposed to integrate spherical harmonics to a convolution. You et al.~\cite{you-prin-aaai20} transform the point cloud onto spherical voxel grids and apply convolution in the transformed domain. 
A great benefit of such techniques is that it allows \emph{consistent} predictions across training/testing scenarios with or without rotations being applied to the data, and they can generalize robustly to inputs with unseen rotations. Despite that, so far these techniques share a common limitation: their performance is inferior to that in translation-invariant point cloud convolution. A typical example is the accuracy in object classification task on ModelNet40 dataset~\cite{wu-3dshapenets-cvpr15}. State-of-the-art techniques such as
PointNet~\cite{qi2017pointnet}, PointNet++~\cite{qi2017pointnet++}, PointCNN~\cite{li2018pointcnn}, or ShellNet~\cite{zhang-shellnet-iccv19} report between 89\% to 93\% of accuracy while techniques with rotation-invariant convolution only report up to 86\% of accuracy~\cite{zhang-riconv-3dv19,poulenard-spherical-3dv19}. Our work in this paper is dedicated to analyze and address this problem.

\section{Background}
\label{ri_feat}

\begin{figure}[t]
	\centering
	\includegraphics[width=\linewidth]{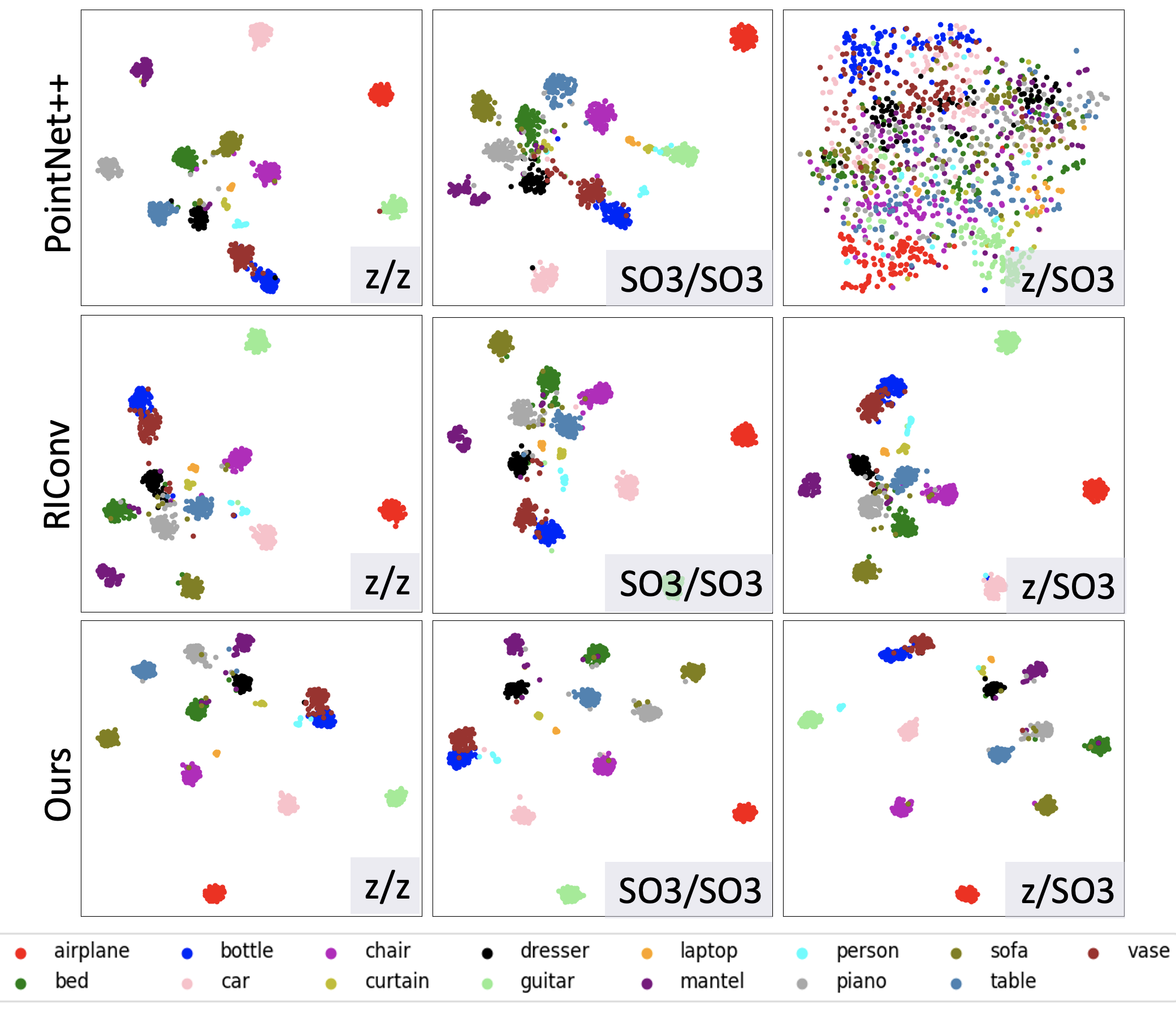}
	\caption{t-SNE comparisons of the latent features for PointNet++~\cite{qi2017pointnet++}, RIConv~\cite{zhang-riconv-3dv19}, and our method under three different rotation settings. The clusters in the t-SNEs show that to make good decisions in object classification, it is desirable to have the cluster boundaries as separated as possible.}
	\label{fig:tsne}
\end{figure}

Let us first analyze the performance of existing point cloud convolutions and their rotation-invariant counterparts. We select object classification task as the key task for our analysis. An observation is that the classification accuracy drops when rotation-invariant convolution is applied. We further dissect this phenomenon by visualizing the latent space learnt by the neural networks using t-SNE~\cite{maaten-tsne-2008}. The results are shown in Figure~\ref{fig:tsne}. 

In this figure, we follow Esteves et al.~\cite{esteves2018learning} and Zhang et al.~\cite{zhang-riconv-3dv19} to evaluate three scenarios for object classification: z/z, SO3/SO3, and z/SO3. In case z/z, we use data augmented with rotation about gravity axis for training and testing. In case SO3/SO3, we use data augmented with arbitrary rotations for training and testing. In case z/SO3, we train with data by z-rotations and test with data by SO3 rotations. The first scenario has been extensively evaluated by previous point cloud convolution methods. The second and third scenario is specially designed to evaluate rotation invariance. The third scenario is the most challenging as it is designed to test whether a convolution can generalize well to unseen rotations. 

As can be seen, latent space learnt by rotation-invariant convolution such as RIConv by Zhang et al.~\cite{zhang-riconv-3dv19} does not exhibit good discrimination among classes. The main difference between such convolution and traditional point cloud convolution is that it no longer works with point coordinates at start. In the case of RIConv, the points are transformed into Euclidean based features including distances and angles, which are not as unique as point coordinates since many points can share the same distance and angles. This is well reflected into the t-SNE in the first column (z/z) in Figure~\ref{fig:tsne}.  PointNet++~\cite{qi2017pointnet++} has a good separation among the clusters while RIConv~\cite{zhang-riconv-3dv19} has more condensed clusters in the center, resulting in more ambiguities during classification. 

Similarly, in the second column (SO3/SO3), PointNet++ and RIConv has similar clustering, which explains their similar performance in the classification (see more quantitative comparisons in Table~\ref{tab_classification_modelnet40}). Finally, the third column (z/SO3) highlights the strength of rotation-invariant convolutions as they can still maintain consistent predictions and generalize well to unseen conditions. In this case, the t-SNEs show that PointNet++ cannot generalize effectively. 

The goal of our work is to devise a convolution that can output highly distinctive rotation-invariant features. Here we achieve this by introducing features from a global context to design a new rotation-invariant convolution. We are inspired by the fact that for each point in a point cloud, its 3D coordinates encode global information. Such global information is lost when one converts the coordinates into some rotation-invariant features such as distance and angles as done by Zhang et al.~\cite{zhang-riconv-3dv19}. 

\section{Our Method}

Our rotation-invariant convolution is built upon two key concepts: a repeatable and robust local reference frame and a global context using anchors. The idea of using local reference frames is related to spatial transformer~\cite{jaderberg-stn-nips15} which is also leveraged by PointNet~\cite{qi2017pointnet}. However, as spatial transformer is data-driven, it does not work well to unseen conditions such as the z/SO3 test in Figure~\ref{fig:tsne}. To achieve robustness, we build local reference frames (LRFs) at the keypoints of the point cloud so that features can be learnt in such local spaces. At a keypoint, not only points in its local neighborhood can strongly affect the construction of the reference frame, but non-neighboring points can also contribute to such construction. It is well known that repeatable and robust LRFs are keys to traditional 3D point descriptors~\cite{tombari-shot-eccv10}. 

After the LRFs are constructed, theoretically we can simply proceed to learn features of the local point sets. However, as previously mentioned, global shape information are also useful for feature learning. We also retain such global information and integrate them into the convolution. Here we achieve this through \emph{anchors}. Each anchor is defined as a representative point in each subspace formed by the axes of the LRF. Given a LRF, it is possible to construct eight subspaces. At each LRF, the anchors thus approximate global features of the point cloud and we integrate such features to define our convolution.

\begin{figure*}[t]
	\centering
	\includegraphics[width=0.9\linewidth]{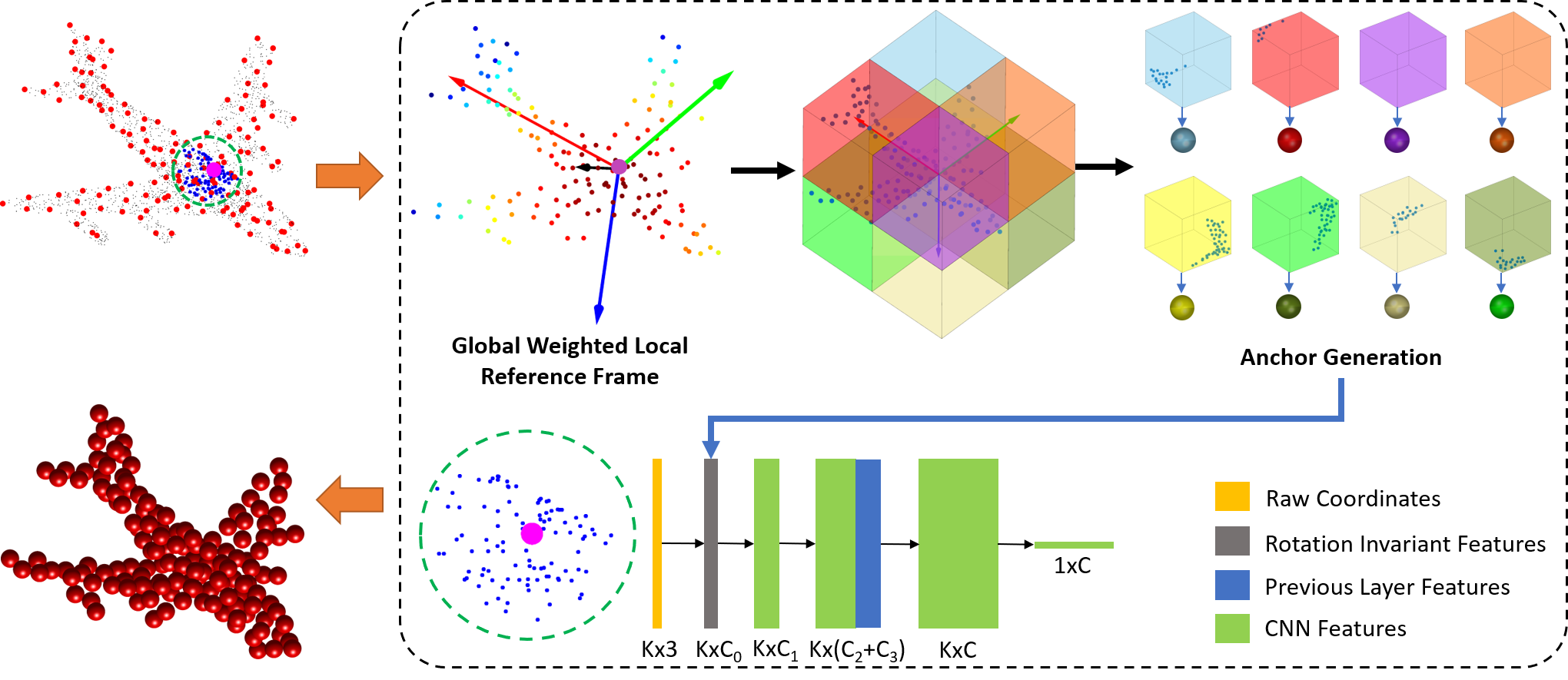}
	\caption{Global context aware convolution (\ourconv) for learning point cloud features comprises of two main steps: (1) Transform into rotation-invariant feature space: for an input point cloud (the upper left plane model), the red dots indicate the keypoints extracted by farthest point sampling. At each keypoint, we first establish a local reference frame (LRF) by employing weights from \emph{all} other keypoints. The 3D coordinates of the keypoint neighbors are projected to the local space spanned by the LRF to obtain rotation invariant features; and (2) Global feature learning with anchors: eight anchors are constructed to represent eight bins that spans the half-spaces due to the LRF. The local-global relation between the points in a neighborhood and the shape approximates in anchors are folded by a 1D convolution to output final rotation-invariant features.}
	\label{fig:conv}
\end{figure*}

\subsection{Globally Weighted Local Reference Frames}
For an input point set, we use farthest point sampling to select a set of keypoints which can fully cover the underlying point cloud and denoted as $Q$. For each keypoint $p\in Q$, we use it as a query to obtain local region $\Omega_p$ centroid at $p$. We wish to use deep learning to extract rotation invariant features from the local region. To begin with local features learning, it is necessary to construct a local reference frame (LRF) such that the 3D coordinates can be transformed into rotation invariant features.  
The unit vectors of the LRF at $p$ can be determined by normalizing the eigenvectors of the covariance matrix
\begin{align}
\Sigma_p = \sum_{i=1}^{N_{sub}} (x_i - p) (x_i - p)^\top,
\end{align}
where $N_{sub}$ is the number of points in the local region and $x_i\in \Omega_p$. 
However, the LRF via such computation is unstable and sensitive to noise. Slight point variations can affect the LRF and make it not repeatable. 
Moreover, when a local region $\Omega_p$ undergoes some rotations, ambiguity can arise, reducing the distinctiveness of the local features. For example, it is hard to tell apart a corner region on a bed and on a floor/wall/ceiling in the presence of arbitrary rotations. To solve these problems, we establish more reliable LRFs by utilizing all query points of $Q$ in the construction: 
\begin{align}
\Sigma_q = \sum_{i=1}^{N} w_i (q_i - p) (q_i - p)^\top,
\end{align}
where $w_i$ is the weight that controls how a point in the point set contributes to the matrix. The weight is defined by
\begin{align}
w_i = \frac{m - \| q_i - p \|}{\sum_{i=1}^{N} m - \| q_i - p \|},
\end{align}
where $m = \max_{i=1..N}(\| q_i - p \|)$. 
Intuitively, this weight allows nearby points of $p$ to have large contributions to the covariance matrix, and thus greatly affect the LRF. Points further away from $p$ however can contribute globally to the robustness of the LRF. Such weighted LRF construction is a fundamental step in 3D hand-crafted features~\cite{tombari-shot-eccv10}, which can be easily integrated into our proposed convolution.

A typical problem in defining LRFs is the sign flipping, i.e., the LRF signs should not vary for the same point set~\cite{tombari-shot-eccv10}. There are multiple ways to resolve the ambiguity; here we disambiguate the signs of the eigenvectors by orienting them to the global vector $O$ defined by
\begin{align}
O = \sum_{i=1}^N w_i (q_i - p),
\end{align} 
which represents the main orientation of the whole model from the perspective of point $p$.

\subsection{Anchor Point Generation}
Theoretically, it is possible to perform convolution on the point set transformed into local coordinates using the constructed LRF. However, it is wasteful to discard global information from the original coordinates as such information can further improve feature distinctiveness. Our idea here is to use anchor points to retain such information in a compact way. 

Specifically, to establish the anchors, we divide the whole input point cloud into eight bins, as shown in Figure~\ref{fig:conv}. In each bin, we use the barycenter of the local point set in that bin as the anchor point. Such anchors are crude approximations to the global input shape, and therefore they convey useful information for the convolution. 

It is worth noting that there are many ways to define anchors in our case. For example, one can choose to use more bins or all the original point coordinates as anchors, but those will significantly increase computation time for the convolution. We empirically use eight bins as it strikes a balance between the amount of global information retained and the running time.  

\subsection{Global Context Aware Convolution}
With the LRFs and anchors points defined, we are now ready to construct our Global Context Aware Convolution (\ourconv) to learn the rotation invariant features. 
Let us consider a point set $P = \{x_i\}$ where $x_i$ represents 3D coordinates of the point $i$. Let $\Omega_i$ be a local point set centered at $x_i$. 
A typical convolution to learn the features of $\Omega_i$ can be written as
\begin{align} 
\mathbf{f}(\Omega_i) = \sigma( \mathcal{A} ( \{ \mathcal{T}(\mathbf{f}_{x_i}) : \forall i \} ) )
\end{align}
This formula indicates that features of each point in the point set are first transformed before being aggregated by the aggregation function $\mathcal{A}$ and passed to an activation function $\sigma$. A popular choice of $\mathcal{A}$ is maxpooling, which supports permutation invariance in the orders of the input point features~\cite{qi2017pointnet}. There are a few ways to define the transformation function $\mathcal{T}$. In PointNet~\cite{qi2017pointnet}, it is defined by
\begin{align}
\mathcal{T}(\mathbf{f}_{x_i}) = \mathbf{w}_i \cdot \mathbf{f}_{x_i}
\end{align}
where $\cdot$ indicates the element-wise product. This product however ignores the contribution of features from neighboring points $x_j$ to center $x_i$. 
To further incorporate such neighbor information, Liu et al.~\cite{liu2019relation} proposed to define the weights by a mapping from a relation vector $
\mathbf{h}_{ij}$ between a point $x_i$ and its neighbor $x_j$.

Here our goal is to define the weights by using the local point set and the anchors. 
We project both the local point set and anchor points onto the LRF system such that the global 3D coordinates are transformed to a local frame: 
\begin{align}
x_i' = LRF(x_i), \quad \quad a_i' = LRF(a_i).
\end{align}
where $x_i$ and $a_i$ represents the global point and anchor, and $x_i'$ and $a_i'$ represents the local point and anchor, respectively. From here, we aim to relate the weights to such coordinates.
Given a pair of a local point $x_i'$ and an anchor $a_j'$, we define their relation as 
\begin{align}
\mathbf{h}(x_i', a_j') = (x_i' - a_j', \| x_i' - a_j' \|)
\end{align} 
which can be represented by a $1 \times 4$ vector. We stack the features over eight anchors into an $8 \times 4$ matrix. 

Our convolution can then be defined as a 1D convolution $\mathcal{K}$ that transforms such matrix into a feature vector. The kernel of the convolution is $1 \times 8$.   
\begin{align}
\mathcal{T}(\mathbf{f}_{\Omega_i}) = \mathbf{w}_{i} \cdot \mathbf{f}_{x_i} = (\mathcal{K} \star \mathbf{h_i}) \cdot \mathbf{f}_{x_i}
\end{align}
Note that in this formula, we operate on local coordinates, and we use the anchors $a_i'$ to approximate features from neighboring points. This allows us to have two main advantages. First, our convolution only needs local features to operate. Second, the LRFs allow that the learnt features are rotation invariant by definition, without the need of data augmentation during training. Our features can generalize easily to unseen rotations, and we also save a lot of computation during training.

\subsection{Network Architecture}
\begin{figure}[t]
	\centering
	\includegraphics[width=\linewidth]{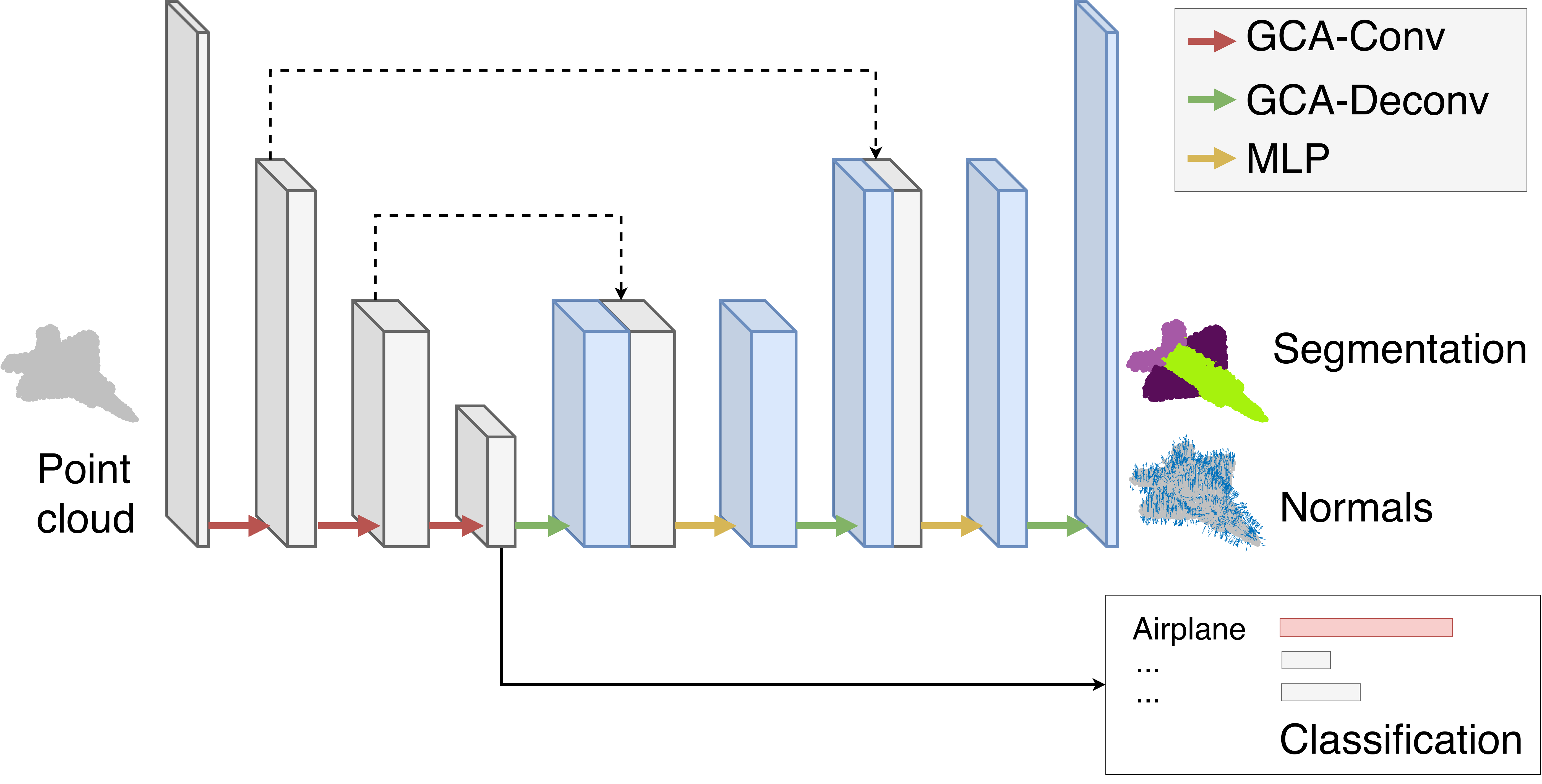}
	\caption{Our network architecture with the proposed point cloud convolution. We use three convolution layers to extract point cloud features before fully connected layers for object classification. We use the same encoder-decoder style architecture with skip connections for object part segmentation and normal estimation task.}
	\label{fig:gcanet}
\end{figure}

We use the proposed convolution to design three neural networks for object classification, object part segmentation, and normals estimation, respectively. The architecture is shown in Figure~\ref{fig:gcanet}. Our classification network has a standard architecture and uses three consecutive layers of convolution (with point downsampling) followed by fully connected layers (256, 128) to output the probability map. In three layers of convolutions, the output channels are set as 128, 256, 512 respectively, and the downsampling numbers are set as 512, 128 and 32 respectively. The neural network for object part segmentation and normal estimation has a decoder branch that includes skip connections and gradually upsamples the point cloud to the original resolution. We use MLP after a skip connection to unify and transform the combined features to have a valid size before deconvolution. Our deconvolution is defined similarly to GCAConv. The minor difference is that it gradually outputs denser points with fewer features.
 
\section{Experimental Results}
\label{experiment}
In this section, we evaluate our method on the 3D object classification, object part segmentation, shape retrieval, and normal estimation task. We implemented our method in TensorFlow~\cite{abadi2016tensorflow}. We use a batch size of $32$ to train object classification and $16$ to train object part segmentation, shape retrieval, and normal estimation. The training is performed with Adam optimizer with an initial learning rate set to 0.001. The experiments are conducted on a machine with an Intel(R) Core(TM) i7-6900K CPU equipped with an NVIDIA GTX TITAN X GPU. 

\subsection{Classification on ModelNet40}
\label{class_modelnet40}

\begin{table*}[t!]
	\centering
	\begin{tabular}{l|lll|ccc|cc}
		\toprule
		Method  & Format & Input size & Params. & z/z & SO3/SO3 & z/SO3 & Average acc. & Acc. std.\\
		\midrule  
		VoxNet \cite{huang2018recurrent} & voxel & $30^{3}$ &0.9M & 83.0 & 87.3 & - & 85.2  & 3.0 \\ 
		SubVolSup \cite{qi2016volumetric}& voxel & $30^{3}$ &17M  & 88.5 & 82.7 & 36.6 & 69.3 & 28.4 \\ 
		Spherical CNN \cite{esteves2018learning}  & voxel & $2\times 64^{2}$ & 0.5M & 88.9 & 86.9 & 78.6 & 84.8 & 5.5\\ 
		MVCNN 80x \cite{su2015multi}  & view & $80\times 224^{2}$ & 99M & 90.2 & 86.0 & 81.5 & 85.9 & 4.3 \\
		PointNet \cite{qi2017pointnet}  & xyz & $1024\times 3$ & 3.5M & 87.0 & 80.3 & 21.6 & 63.0 & 41.0 \\ 
		PointNet++ \cite{qi2017pointnet++}  & xyz & $1024\times 3$ & 1.4M & 89.3 & 85.0 & 28.6 & 67.6 & 33.8\\
		PointCNN \cite{li2018pointcnn}  & xyz & $1024\times 3$ & 0.60M & 91.3 & 84.5 & 41.2 & 72.3 & 27.2 \\
		RS-CNN \cite{liu2019relation}  & xyz & $1024\times 3$ & 1.41M & 90.3 & 82.6 & 48.7 & 73.9 & 22.1\\
		\midrule
		RIConv ~\cite{zhang-riconv-3dv19} & xyz & 1024 $\times 3$ & 0.70M & 86.5 & 86.4 & 86.4 & 86.4 & 0.1\\
		SPHNet ~\cite{poulenard-spherical-3dv19} & xyz & 1024 $\times 3$ &  2.9M & 87.0 & 87.6 & 86.6 & 87.1 & 0.5 \\
		SFCNN~\cite{rao-spherical-cvpr19} & xyz & 1024 $\times 3$ &  - & \textbf{91.4} & \textbf{90.1} & 84.8 & 88.8 & 3.5 \\
		ClusterNet ~\cite{chen2019clusternet} & xyz & 1024 $\times 3$ &  - & 87.1 & 87.1 & 87.1 & 87.1 & \textbf{0.0} \\
		\midrule 
		Ours (w/o anchor) & xyz & 1024 $\times 3$ & 0.21M & 86.3 & 86.2 & 86.2 & 86.2 & \textbf{0.0}\\
		Ours & xyz & 1024 $\times 3$ & 0.39M & 89.0 & 89.2 & \textbf{89.1} & \textbf{89.1} & \textbf{0.0}\\
		
		\bottomrule
	\end{tabular}
	\caption{Comparisons of the classification accuracy (\%) on the ModelNet40 dataset. On average, our method has the best accuracy and lowest accuracy deviation in all cases.}
	\label{tab_classification_modelnet40}
\end{table*}

Object classification is the main task in our evaluation. We train the classification network by using the ModelNet40 variant of the ModelNet dataset~\cite{wu20153d}. ModelNet40 contains CAD models from 40 categories such as airplane, bottle, chair, dresser, vase, etc. We use the preprocessed data from PointNet~\cite{qi2017pointnet} that consists of $9,843$ models for training and $2,468$ models for testing. 
We use point clouds of size 1024 in this task. Each point is represented by $(x,y,z)$ coordinates in the Euclidean space. The training takes approximately 11 hours to converge in 250 epochs. 

Following Esteves et al.~\cite{esteves2018learning} and Zhang et al.~\cite{zhang-riconv-3dv19}, we evaluate the performance of object classification with three scenarios: (1) using data augmented with rotation about gravity axis (z/z) for training and testing, (2) using data augmented with arbitrary rotations (SO3/SO3) for training and testing, and (3) training with data by z-rotations and testing with data by SO3 rotations (z/SO3). It is expected that rotation-invariant convolutions should work well in the z/SO3 scenario. 

Table~\ref{tab_classification_modelnet40} details the results of this experiment, which confirms the effectiveness of the proposed rotation-invariant convolution. 
As can be seen, on average, not only our classification accuracy outperforms the state-of-the-art translation-invariant point cloud convolution, the performance is also consistent across three scenarios. For rotation-invariant convolutions, our method outperforms the accuracy of RIConv~\cite{zhang-riconv-3dv19}, SPHNet~\cite{poulenard-spherical-3dv19}, and ClusterNet~\cite{chen2019clusternet} by a good margin. Our method is slightly more accurate than SFCNN~\cite{rao-spherical-cvpr19} but much more consistent.


\subsubsection{Ablation Studies}

\paragraph{Network Design.}

\newcommand{\tabincell}[2]{\begin{tabular}{@{}#1@{}}#2\vspace{-0.15in}\end{tabular}}
\begin{table}[b]
	\centering
	\begin{tabularx}{\linewidth}{Y|YYYYY}
		\toprule
		Model  & Weight & $O$ Vector & Anchor &Rot. Aug. & Acc. \\
		\midrule
		
		\tabincell{c}{A \\B \\ C \\ D \\ E} 

		& \tabincell{c}{\checkmark \\  \\  \\ \checkmark \\ \checkmark} 
		
		& \tabincell{c}{\checkmark \\ \checkmark \\   \\ \checkmark \\ \checkmark} 
		
		& \tabincell{c}{\checkmark \\ \checkmark \\ \checkmark \\   \\ \checkmark} 
		
		& \tabincell{c}{\checkmark \\ \checkmark \\ \checkmark \\  \checkmark \\  } 
		
		& \tabincell{c}{89.2\\ 87.1 \\ 86.7 \\ 86.6 \\ 89.2}\\ 
		
		\bottomrule
	\end{tabularx}
	\caption{An evaluation of our network design. It shows that weighted LRF, resolving LRF sign ambiguity, and global anchor play an important role for good performance.}
	\label{tab_ablation}
\end{table}

We conduct an ablation study on the ModelNet40 dataset for the classification task (Table~\ref{tab_ablation}). We examine four settings in our convolution: (1) the globally weighted LRFs with main orientation (Weight), (2) the use of main orientation to resolve the LRF sign ambiguity ($O$ vector), (3) the use of anchors for global context (Anchor), and (4) the data augmentation with rotations used for the training (Rot. Aug.). Five models (A-E) are used to study the effects of these settings by turning them on/off.

Model A is our baseline setting with all settings on. Model B tests the importance of the weights for computing LRFs and the main orientation. It can be seen that without such weights, the accuracy decreases to 87.1\%. The main reason is that the LRFs and the main orientation are more noisy and less repeatable in such case. 
Next, in model C we further turn off the $O$ vector to test the stability of the LRFs without sign correction. The accuracy further decreases to 86.7\%. This verifies that constructing stable LRFs is key to good network performance. In model D, we turn off the global anchor. In this case, only the local points are used for feature extraction. Thanks to the LRFs, the local features are still effective despite of mild accuracy drop. In model E, we test the performance without rotation augmentation scheme during the training procedure. We find the accuracy is not affected by data augmentation as \ourconv already achieves exact rotation invariance.

\begin{table}[t]
	\begin{center}
		\begin{tabular}{c|cccc}
			\toprule
			Number of Anchors   & 1 & 2 & 4 & 8 \\
			\midrule
			Accuracy     & 87.3   &87.8   & 88.5  & 89.2  \\
			\bottomrule
		\end{tabular}
	\end{center}
	\caption{Classification accuracy (\%) on ModelNet40~\cite{wu20153d} with different number of anchors.}
	\label{tab_anchors}
\end{table}

\paragraph{Comparison to learned LRFs.}
It is generally tempting to learn the LRFs to design rotation-invariant convolution. Here we compare this method to our proposed LRFs. We use a two-layer MLP to predict the LRFs and then use them to transform the input point coordinates into a local coordinates before proceeding for convolution as described in the main paper. We found that predicting LRFs works well in z/z and SO3/SO3 mode, with both scenarios achieved accuracies of 89.3\% and 89.2\%, respectively. However, using data-driven LRFs makes the convolution only \emph{rotation-aware}, but not exactly rotation-invariant. Such convolution fails to generalize to unseen rotations in the z/SO3 scenario with the accuracy of 36.2\%.

\paragraph{Number of Anchors.}
From the ablation studies, we see that without global anchors, the performance is decreased. Here, we further analyze the effects of the number of anchors by investigating the performance on ModelNet40 with a different number of anchors. The qualitative results are shown in Table~\ref{tab_anchors}. We can see that with only one anchor, the accuracy decreases to 87.3\%, but still higher than RIConv which is around 86.4\%. This shows the advantages of global information. With the number goes on, the accuracy also increases. We empirically use eight anchors as it strikes a balance between the amount of global information retained and the running time.

\subsection{Object Part Segmentation on ShapeNet}
\label{seg_shapenet}

In addition to object classification, we evaluate our method to output a label for each point in the point cloud, resulting in object part segmentation. We use the 3D models in ShapeNet~\cite{chang2015shapenet} to train our network with point size of 2048 in this task. It takes roughly 36 hours for the training to complete 300 epochs.

\begin{figure}[t]
	\centering
	\includegraphics[width=0.9\linewidth]{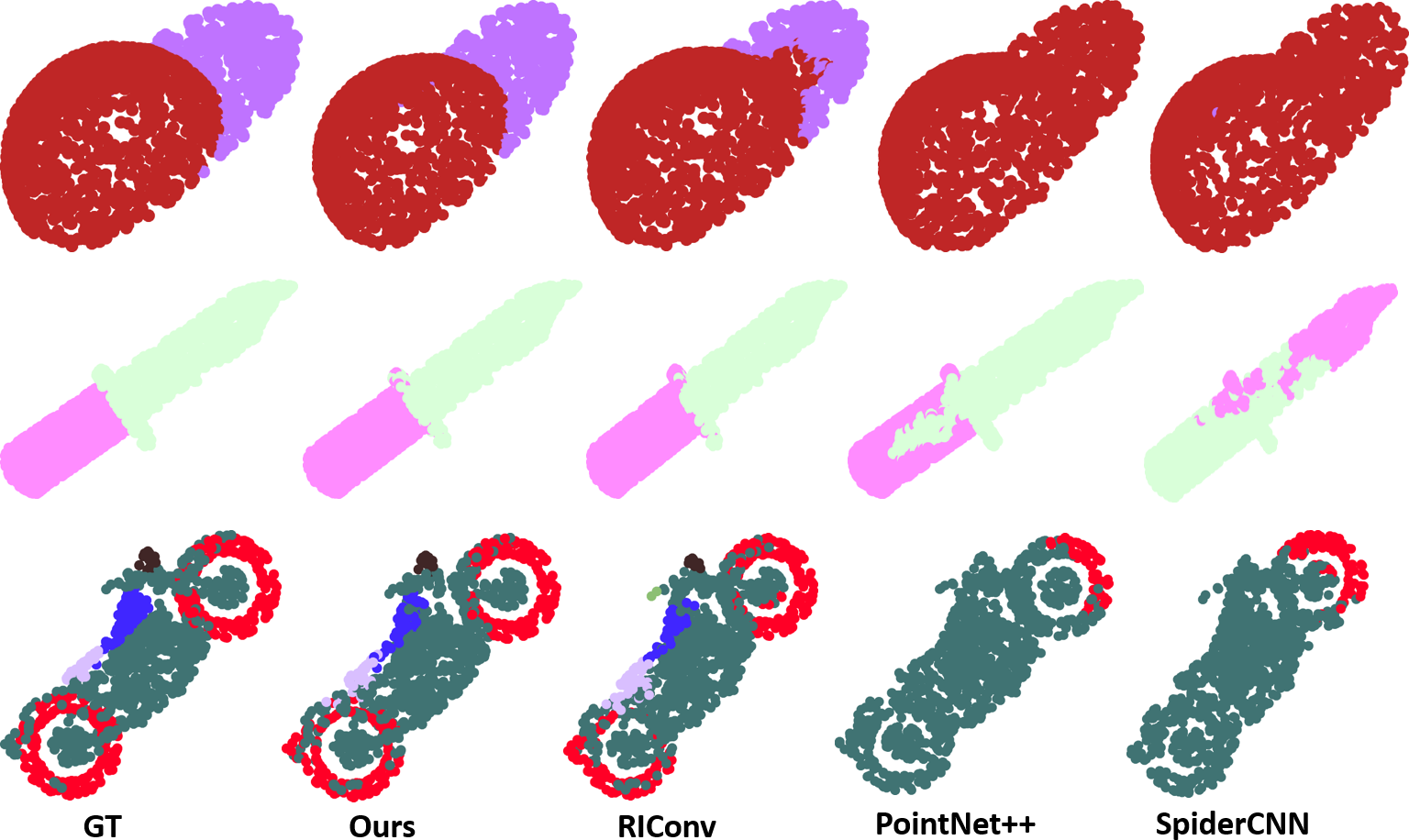}
	\caption{Qualitative comparisons of part segmentation for \ourconv, RIConv ~\cite{zhang-riconv-3dv19}, PointNet++~\cite{qi2017pointnet++}, SpiderCNN \cite{xu2018spidercnn} under the z/SO3 rotation mode (from the left column to the right column).}
	\label{fig:part_seg}
\end{figure}

The quantitative and qualitative results are shown in Table~\ref{tab_segmentation} and Figure~\ref{fig:part_seg}, respectively. 
In this task, we achieve start-of-the-art results for both SO3/SO3 and z/SO3 scenarios. Our method outperforms RIConv~\cite{zhang-riconv-3dv19} by almost $2\%$ of accuracy. 
From Figure~\ref{fig:part_seg}, we can clearly see that with z/SO3 mode methods like PointNet++ and SpiderCNN can not work well. This is easy to explain as these methods use the raw xyz coordinates as input for training, thus cannot well understand unknown rotations. RIConv~\cite{zhang-riconv-3dv19} works better as it converts xyz coordinates into rotation invariant format like distances and angles before training. However, it still has difficulties in recognizing the boundaries while our method can treat these regions well by incorporating global context information (see column 2 and 3 in Figure~\ref{fig:part_seg}).

\begin{table}[t]
	\begin{center}
		\begin{tabular}{l|l|cc}
			\toprule
			Method   & input & SO3/SO3 & z/SO3  \\
			\midrule
			PointNet \cite{qi2017pointnet}     & xyz        & 74.4   & 37.8  \\
			PointNet++ \cite{qi2017pointnet++} & xyz+normal & 76.7   & 48.2  \\
			PointCNN \cite{li2018pointcnn}     & xyz        & 71.4   & 34.7  \\
			DGCNN \cite{wang2018edgeconv}      & xyz        & 73.3   & 37.4  \\
			SpiderCNN \cite{xu2018spidercnn}   & xyz+normal & 72.3   & 42.9  \\
			RS-CNN \cite{liu2019relation}      & xyz        & 72.5   & 36.5  \\
			RIConv ~\cite{zhang-riconv-3dv19}  & xyz        & 75.5   & 75.3  \\
			
			\midrule
			Ours (w/o anchor)                  & xyz &  73.2 &  73.6 \\
			Ours                               & xyz &  \textbf{77.3} &  \textbf{77.2} \\
			\bottomrule
		\end{tabular}
	\end{center}
	\caption{Comparisons of object part segmentation performed on ShapeNet dataset~\cite{chang2015shapenet}. The mean per-class IoU (mIoU, \%) is used to measure the accuracy under two challenging rotation modes: SO3/SO3 and z/SO3.}
	\label{tab_segmentation}
\end{table}

\subsection{Shape Retrieval}
\label{retrieval}
A popular evaluation of rotation invariance on 3D shape is the shape retrieval task~\cite{savva2016shrec16}. Here we conducted experiments on ShapeNet Core~\cite{wu20153d}, following the perturbed protocol of the
SHREC’17 3D shape retrieval contest~\cite{savva2016shrec16} and the experiment setting of SFCNN~\cite{rao-spherical-cvpr19}. 
We use the same output features from the bottleneck layer in the network (similar to features used in the classification task; see Figure~\ref{fig:gcanet}).  
We compare with methods proposed in SHREC’17~\cite{furuya2016deep,tatsuma2009multi,bai2016gift} and two recent methods on rotation-invariant convolution~\cite{esteves2018learning,rao-spherical-cvpr19}. The results are shown in Table~\ref{tab_retrieval}.
It can be seen that our method achieves the state-of-the-art accuracy, outperforming previous methods for most evaluation metrics.

\begin{table*}[t]
	\centering
	\begin{tabular}{l|ccccc|ccccc|c}
		\toprule
		&   &     &micro &   &      &   &    &macro&    &     & \\
		Method &PN &R@N  &F1@N &mAP & NDCG &PN &R@N &F1@N &mAP &NDCG & Score\\
		\midrule  
		Furuya~\cite{furuya2016deep}  &81.4 &68.3 &70.6 &65.6 &75.4 &60.7 &53.9 &50.3 &47.6 &56.0 &56.6\\
		Tatsuma~\cite{tatsuma2009multi}  &70.5 &\textbf{76.9} &71.9 &69.6 &78.3 &42.4 &\textbf{56.3} &43.4 &41.8 &47.9 &55.7 \\
		Zhou~\cite{bai2016gift}   &66.0 &65.0 &64.3 &56.7 &70.1 &44.3 &50.8 &43.7 &40.6 &51.3 &48.7\\
		\midrule
		Spherical CNN~\cite{esteves2018learning}  &71.7 &73.7 &- &68.5 &- &45.0 &55.0 &- &44.4 &- &56.5\\
		SFCNN~\cite{rao-spherical-cvpr19}  &77.8 &75.1 &\textbf{75.2} &70.5 &\textbf{81.3} &65.6 &53.9 &\textbf{53.6} &48.3 &58.0 &59.4\\
		\midrule 
		Ours &\textbf{82.9} &76.3 &74.8 &\textbf{70.8} &\textbf{81.3} &\textbf{66.8} &55.9 &51.2 &\textbf{49.0}	&\textbf{58.2}	&\textbf{61.2}\\
		\bottomrule
	\end{tabular}
	\caption{Comparisons of 3D shape retrieval on the ShapeNet Core~\cite{wu20153d}. The accuracy (\%) is reported based on the standard evaluation metrics including precision, recall, f-score, mean average precision (mAP) and normalized discounted cumulative gain (NDCG).}
	\label{tab_retrieval}
\end{table*}

\subsection{Normals Estimation}
\label{normal_estimation}
Normals estimation for point clouds is instrumental in many applications such as point cloud rendering, feature extraction, and surface reconstruction. Here we conduct normals estimation on point clouds using the ModelNet40 dataset. For each model, we uniformly sample $1024$ points from the original data for training. We compute a loss based on the cosines between the predicted unit vectors and the ground truth normals to guide the training. Our results are shown in Table~\ref{tab_normal_modelnet40}. 

\begin{table}[t]
	\centering
	\begin{tabular}{l|ccc|c}
		\toprule
		Method    & z/z & SO3/SO3 & z/SO3 & Err. std.\\
		\midrule  
		PointNet++ \cite{qi2017pointnet++}  & 0.34 & 0.55 & 0.81 & 0.24\\
		RS-CNN \cite{liu2019relation}   & \textbf{0.26} & 0.50 & 0.83 & 0.29 \\
		RIConv ~\cite{zhang-riconv-3dv19}   & 1.33 & 1.30 & 1.30 & 0.02\\
		\midrule 
		Ours & 0.42 &  \textbf{0.42} &  \textbf{0.44} & \textbf{0.01}\\
		\bottomrule
	\end{tabular}
	\caption{Comparisons of the normal estimation on ModelNet40. The accuracy is reported on three test cases: training and testing with z/z, SO3/SO3 and z/SO3 rotation, respectively. Our method has good accuracy and lowest accuracy deviation in all cases.}
	\label{tab_normal_modelnet40}
\end{table}

\begin{figure}[t]
	\centering
	\includegraphics[width=0.9\linewidth]{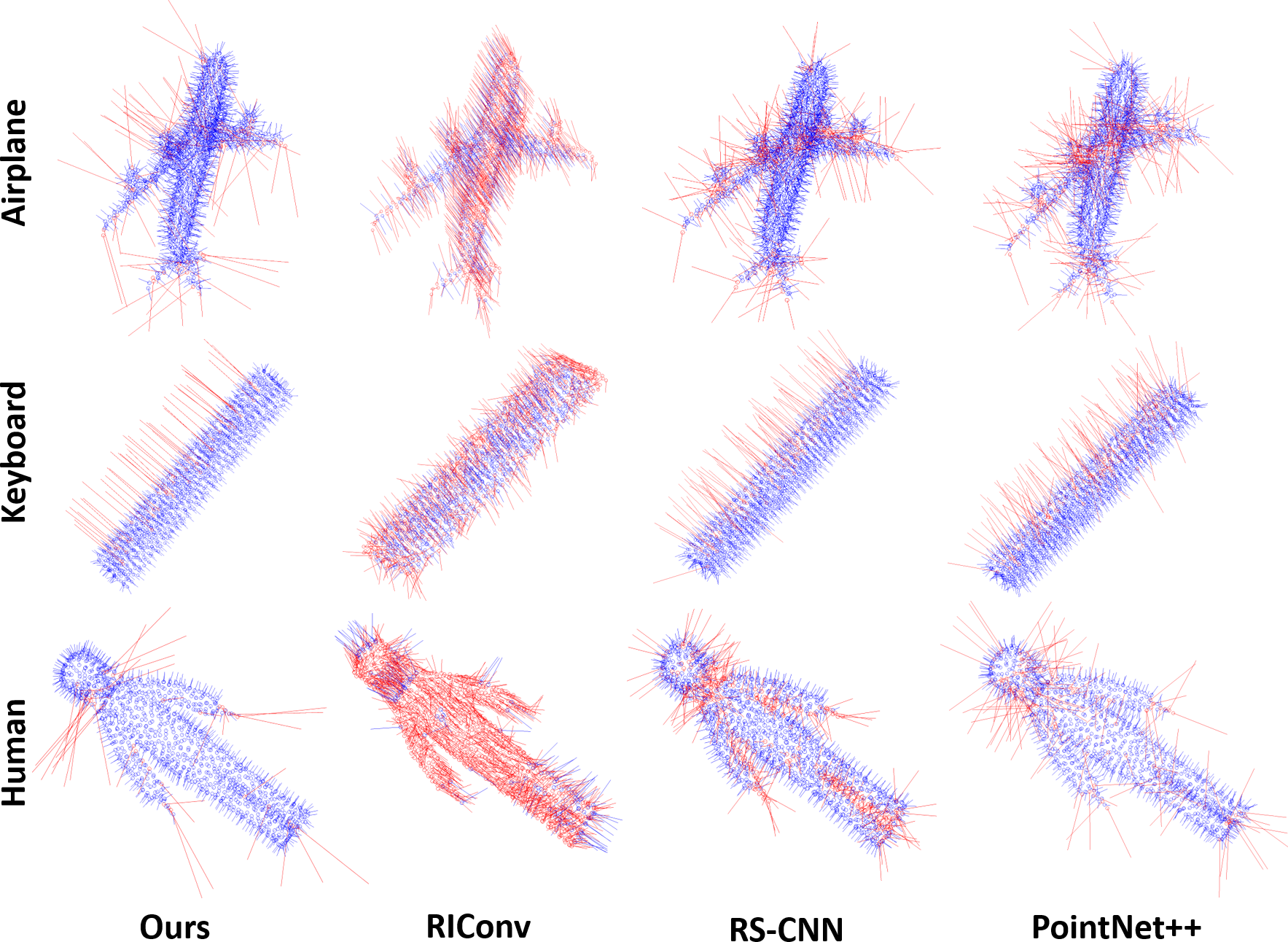}
	\caption{Qualitative comparisons of normal estimation for \ourconv, RIConv ~\cite{zhang-riconv-3dv19}, RS-CNN \cite{liu2019relation}, and PointNet++~\cite{qi2017pointnet++} under the z/SO3 rotation mode (from the left column to the right column).}
	\label{fig:normal}
\end{figure}

In this table, our method achieves the best consistency in predicting normals across three test scenarios. In SO3/SO3 and z/SO3 case, our method is the most accurate. It outperforms other methods by a wide margin. The predicted normals are depicted in Figure~\ref{fig:normal}. We quantize the errors by calculating the angles between the predicted and ground truth normals. In Figure~\ref{fig:normal}, the blue and red vectors depict normals with less than $30^{\circ}$ and greater than $90^{\circ}$ of error. It can be seen that our method is the most accurate visually. It is worth noting that RIConv~\cite{zhang-riconv-3dv19} performs poorly in the normals estimation task because it uses rotation-invariant features that discard the reference coordinate frames, and so the normals of RIConv is not globally consistent.

\section{Conclusion}
\label{conclusion}
In this work, we introduced a novel approach to design rotation-invariant convolution for 3D point clouds. We show that building robust and repeatable local reference frames is critical to boosting the performance of rotation-invariant object classification. In this task, our newly proposed convolution can match the performance of state-of-the-art translation-invariant convolutions. Our work opens up opportunities to narrow down the performance gap between rotation-invariant and translation-invariant convolution in general 3D deep learning, making robust convolutions for 3D point clouds feasible. 

Here we detail a few potential ideas for future research. First, while our proposed method achieves good performance, it is not clear whether local reference frames can be set robustly by a neural network. There is a recent work~\cite{zhu-lrfnet-ar20} that attempts to solve this problem, but the performance on object classification needs further investigation. Second, generalizing point cloud convolutions and object classification to support non-rigid transformations and deformable objects could further improve overall robustness. Finally, more thorough benchmarking rotation-invariant convolutions with real-world data~\cite{uy-scanobjectnn-iccv19} is necessary to understand the impact of such data on the learning of rotation-invariant features. 

{\small
\bibliographystyle{ieee}
\bibliography{egbib}
}

\newpage
\appendix
\renewcommand*\appendixpagename{\Large Supplementary Materials}
\appendixpage

\section{Learning based LRFs}
\subsection{Baseline 1: Predicting LRFs}
As mentioned in the main text, it could be tempting to learn the LRFs to design rotation-aware convolution. For completeness, here we discuss this baseline again. We use a two-layer MLP to predict the LRFs and then use them to transform the input point coordinates into a local coordinates before proceeding for convolution as described in the main paper. We found that predicting LRFs works well in z/z and SO3/SO3 mode, with both scenarios achieved accuracies of 89.3\% and 89.2\%, respectively. However, using data-driven LRFs makes the convolution only rotation-aware, but not exactly rotation-invariant. Such convolution fails to generalize to unseen rotations in the z/SO3 scenario with accuracy 36.2\%. 

\subsection{Baseline 2: Pooling with Sign-Ambiguous LRFs}
Taking the insight from Baseline 1, we proceed to only resolve the ambiguity in constructing the LRFs using learning while using the covariance matrices and their eigenvectors to determine the LRF axes. Here the signs of the LRFs axes are not determined, and instead of resolving this ambiguity as what described in the main paper, here we establish all eight candidates of the LRFs and perform feature learning with all such candidates. The final output features are pooled from the features of each individual candidate. We call this convolution in this baseline the Pooling Convolution (PoolConv).

\begin{figure*}[htb]
	\centering
	\includegraphics[width=0.8\linewidth]{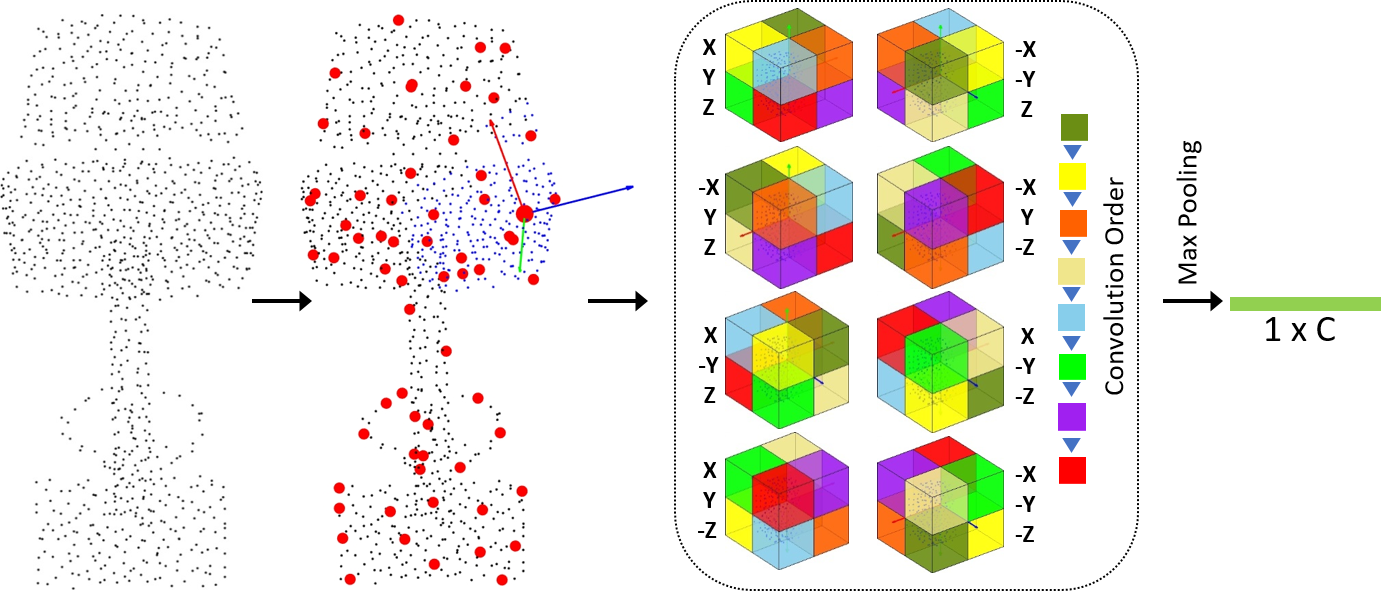}
	\caption{Orientation Pooling Conv.}
	\label{fig:supp_ori_pool}
\end{figure*}

More illustrations can be found in Figure~\ref{fig:supp_ori_pool}. In general, PoolConv can produce the same accuracy (89.1\%) as our method but it has much higher computation. We measure network complexity by the number of trainable parameters, floating point operations (FLOPs), and running time to analyze the network efficiency. With batch size 16, point cloud size 1024 from the ModelNet40 dataset, we report the statistics in Table~\ref{tab:parameters}. Given the minor performance difference but significantly more parameters and training time, PoolConv is not as efficient as our proposed method.

\section{Repeatability}
\begin{figure*}[t]
	\centering
	\includegraphics[width=0.7\linewidth]{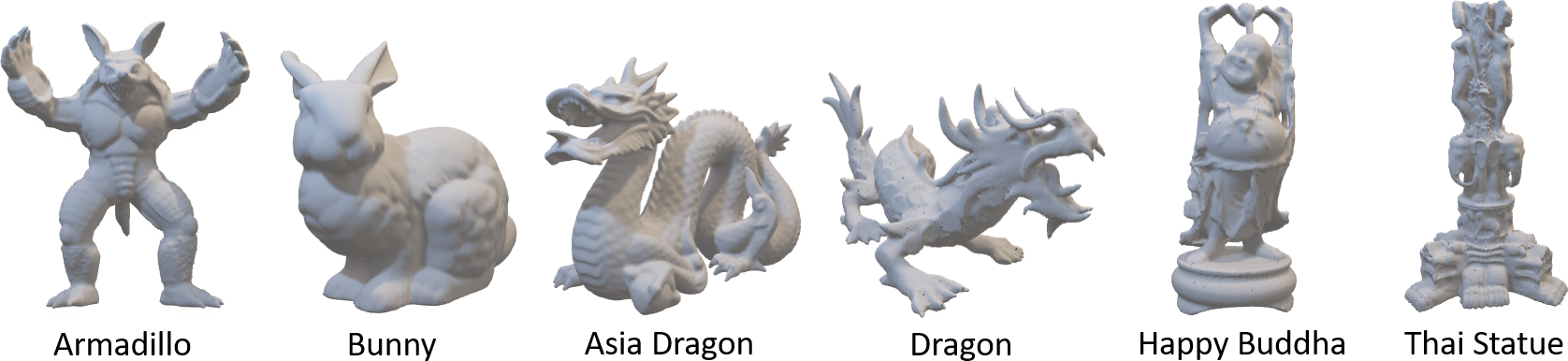}
	\caption{Six models from the Stanford 3D Scanning Repository~\cite{curless1996volumetric}.}
	\label{fig:supp_models}
\end{figure*}

We further clarify the repeatability of the LRFs as it serves as the backbone for our feature learning. We follow Guo et al.~\cite{guo2013rotational} to conduct this experiment (see their section 3.3). Noted that there are also methods that solve LRFs for mesh such as MeshHog~\cite{zaharescu2009surface} and RoPS~\cite{guo2013rotational}. In this study we assume no normal vectors or triangle faces so we omit such methods in our comparison.
We use six models from the Stanford 3D Scanning Repository~\cite{curless1996volumetric} (Figure~\ref{fig:supp_models}). The scenes are created by resampling the models down to 1/2 of their original mesh resolution with Gaussian noise added (0.1 mesh resolution).

\begin{table}[t]
	\begin{center}
		\begin{tabular}{l|p{1cm} p{2.2cm} p{2cm}}
			\toprule
			Method & Params & FLOPs & Time \\
			&  & (Train / Infer) & (Train / Infer) \\
			\midrule
			PoolConv & 0.40M & 116.3B / 12.8B & 0.66s / 0.38s \\
			Ours & \textbf{0.39M} & \textbf{11.0B} / \textbf{1.3B} & \textbf{0.21s} / \textbf{0.16s} \\
			\bottomrule
		\end{tabular}
	\end{center}
	\caption{Comparisons to Baseline 2.}
	\label{tab:parameters}
\end{table}


\begin{figure*}[t!]
	\centering
	\includegraphics[width=\linewidth]{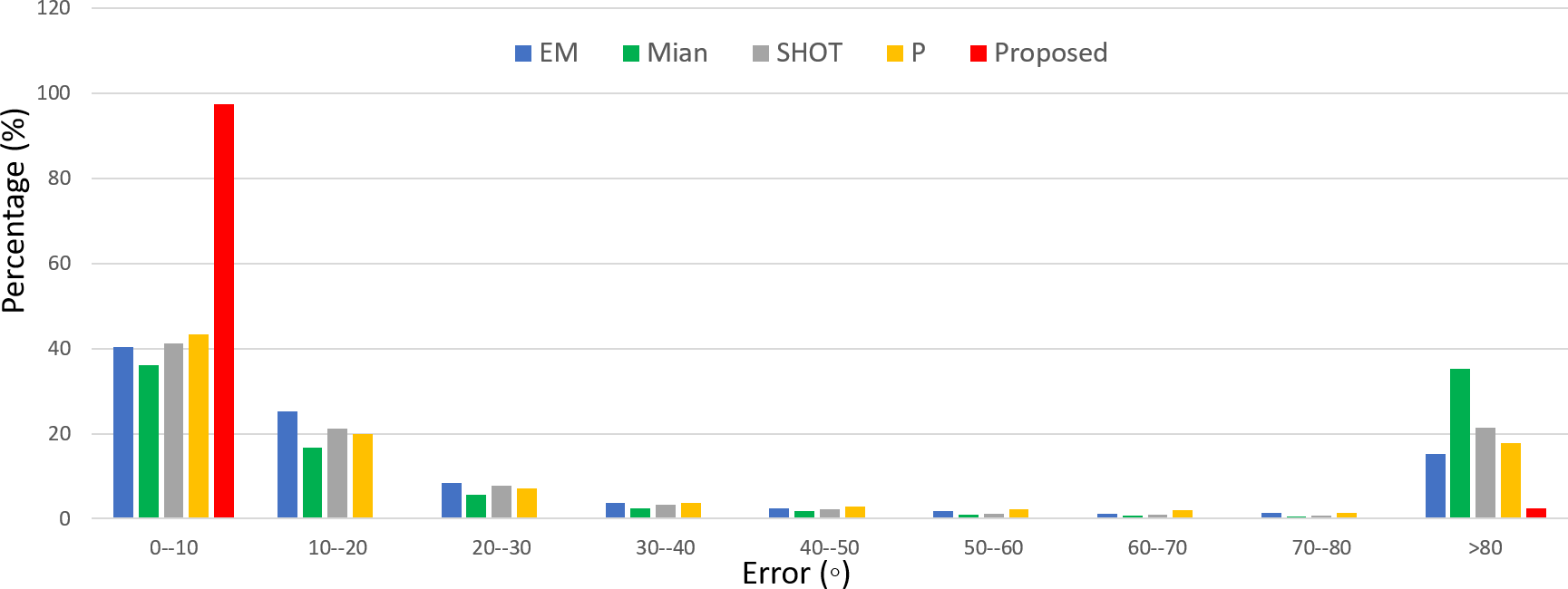}
	\caption{Histogram comparisons of the LRF errors.}
	\label{fig:supp_histogram}
\end{figure*}

From each model, 1000 points are randomly selected and their correspondences in the scene are obtained by searching the closest point in the Euclidean space. Let's denote the pair of points as $(p_{si}, p_{mi})$ from scene and model respectively. The LRFs for these two points are computed as $LRF_{s_i}$ and $LRF_{m_i}$. To measure the similarity between $LRF_{s_i}$ and $LRF_{m_i}$, we use the error evaluation metric provided by Mian et al.~\cite{mian2010repeatability}:
\begin{align}
e_{i} = \arccos\left(\frac{\mathrm{Tr}(LRF_{s_i}LRF_{m_i})-1}{2}\right)\frac{180}{\pi}.
\end{align}
Ideally, $e_{i}$ is zero when there is no error. 
We compare with four existing methods: EM~\cite{novatnack2008scale}, Mian~\cite{mian2010repeatability}, SHOT~\cite{tombari-shot-eccv10}, and P~\cite{petrelli2011repeatability}.
The results are shown in Figure~\ref{fig:supp_histogram}, where the horizontal axis indicates the angular error range and the vertical axis represents the percentage of points. The more points fall into left lower error range, the better of the methods. 
As can be seen, our proposed LRFs have much more low-range angular errors than other methods, but has significantly less high-range errors. This means that our LRFs varies more slowly, and thus allows more consistent predictions.

\section{Per-Class Accuracies}
To further demonstrate the advantages of our proposed convolution operator, we show the per-class accuracies for both classification and part segmentation tasks in this section. 
\subsection{Per-Class Accuracies for Object Classification on ModelNet40}
The per-class accuracies for object classification on ModelNet40 under z/SO3 scenario is shown in Table~\ref{tab:modelnet40_perclass}. Our method outperforms previous methods significantly (ranking 1st in 32 out of 40 classes). 
\begin{table*}
	\centering
	\begin{tabular}{l| p{28pt}p{28pt}ccccp{30pt}p{28pt}}
		\toprule
		Network  & aero & bathtub & bed  & bench & bookshelf & bottle & bowl  & car  \\ 
		\midrule  
		PointNet \cite{qi2017pointnet}  &12.0 &2.0 & 8.0 & 10.0 & 15.0 &14.0&5.0 &12.0\\ 
		
		PointNet++ \cite{qi2017pointnet++} & 53.0  &  2.0  &  18.0  &  10.0  &  29.0  &  22.0  & 20.0  &  13.0  \\ 
		PointCNN \cite{li2018pointcnn} & 60.0 & 10.0 & 20.0 & 10.0 & 20.0 & 37.0 & 25.0 & 34.0   \\ 
		
		RIConv \cite{zhang-riconv-3dv19} &100.0 &82.0 &94.0 &80.0 &93.0 &94.0 &100.0 &98.0 \\
		Ours & \textbf{100.0} & \textbf{90.0} & \textbf{98.0} &\textbf{80.0} & \textbf{95.0} &\textbf{97.0} & \textbf{100.0} & \textbf{98.0} \\
		\midrule 
		& chair & cone & cup & curtain & desk & door & dresser & flower pot\\
		\midrule
		PointNet\cite{qi2017pointnet}  &9.0&15.0 &0.0&0.0&16.3&5.0&8.1&0.0\\
		PointNet++ \cite{qi2017pointnet++}  &  32.0  &  20.0  &  15.0  &  45.0  & 2.3  & 30.0  &  9.3  & 15.0    \\
		PointCNN \cite{li2018pointcnn}  & 46.0  & 25.0  & 15.0  & 40.0 & 34.9 & 30.0 & 32.6 & 25.0 \\  
		
		RIConv \cite{zhang-riconv-3dv19} &96.0 &90.0 & \textbf{60.0} & 95.0 & 79.1 & \textbf{85.0} & \textbf{73.3} & \textbf{30.0}  \\
		Ours & \textbf{98.0} & \textbf{90.0} & 55.0 &\textbf{95.0} & \textbf{81.4} &80.0 & 68.6 & 10.0 \\
		\midrule 
		& glass box & guitar & keyboard & lamp & laptop & mantel & monitor & night stand\\
		\midrule
		PointNet \cite{qi2017pointnet}   &4.0&36.0&5.0&15.0 &15.0&4.0&11.0&3.5\\
		PointNet++ \cite{qi2017pointnet++}  & 11.0  & 47.0  & 50.0  & 10.0    & 15.0 &  10.0  & 36.0  &  1.2  \\
		PointCNN \cite{li2018pointcnn}  & 35.0 & 46.0 & 50.0 &  20.0  & 20.0 & 38.0 & 35.0 & 40.7\\ 
		RIConv \cite{zhang-riconv-3dv19} &96.0 &99.0 &95.0 & 80.0 &95.0 & 91.9 & 97.0 & \textbf{77.9} \\
		Ours & \textbf{97.0} & \textbf{100.0} & \textbf{95.0} &\textbf{85.0} & \textbf{100.0} &\textbf{93.0} & \textbf{98.0} & 73.3 \\
		\midrule 
		 & person & piano & plant & radio & range hood & sink & sofa & stairs \\
		\midrule 
		PointNet \cite{qi2017pointnet}    &5.0&36.7&55.0&5.0&4.0&20.0&11.0&25.0 \\
		PointNet++ \cite{qi2017pointnet++}   & 20.0  &  5.0  & 71.0 &  20.0   & 9.0  &  5.0 & 21.0  & 10.0 \\
		PointCNN \cite{li2018pointcnn}  & 15.0 & 34.0 & 26.0  & 10.0 &  28.0  & 20.0   & 32.0 &  30.0\\ 
		RIConv \cite{zhang-riconv-3dv19} &85.0 &90.8 & 83.0 &55.0 & \textbf{87.0} &\textbf{75.0} & 92.0 &\textbf{85.0} \\
		Ours & \textbf{90.0} & \textbf{91.0} & \textbf{93.0} &\textbf{65.0} & 86.0 &70.0 & \textbf{93.0} & 80.0 \\
		\midrule 
		& stool & table & tent & toilet & tv stand & vase & wardrobe & xbox\\
		\midrule 
		PointNet \cite{qi2017pointnet} &5.0&3.0&5.0&20.0&4.0&26.3&0.0&10.0 \\
		PointNet++ \cite{qi2017pointnet++}  & 10.0  &  9.0  & 15.0  & 13.0 & 2.0  & \textbf{85.0} &15.0 &  20.0 \\
		PointCNN \cite{li2018pointcnn}	 &  20.0 &  36.0 &  15.0 &  33.0  & 29.0 & 70.0   &40.0 &  15.0\\
		RIConv \cite{zhang-riconv-3dv19} &60.0 &80.0 &70.0 &95.0 &78.0 &76.8 &70.0 &65.0\\
		Ours & \textbf{75.0} & \textbf{84.0} & \textbf{95.0} &\textbf{99.0} & \textbf{81.0} &77.0 & \textbf{70.0} & \textbf{75.0} \\
		\bottomrule
	\end{tabular}
	\caption{Per-class accuracy of object classification in z/SO3 scenario with the ModelNet40 dataset~\cite{wu20153d}.}
	\label{tab:modelnet40_perclass}
\end{table*}

\subsection{Per-Class Accuracies for Part Segmentation on ShapeNet}
Here, we also show the per-class accuracies for part segmentation under the SO3/SO3 and z/SO3 scenarios in Table~\ref{tab:objectpart_perclass_so3_so3} and Table~\ref{objectpart_perclass_z_so3} respectively. 
\begin{table*}[t]
	\centering
	\begin{tabular}{p{80pt} p{30pt}p{30pt}p{30pt}p{30pt}p{30pt}p{30pt}p{30pt}p{30pt}p{30pt}p{30pt}p{30pt}p{30pt}p{30pt}p{30pt}p{30pt}p{30pt}}
		\toprule
		Network   & aero & bag & cap & car & chair & earph. & guitar & knife \\ 
		\midrule 
		PointNet \cite{qi2017pointnet}  & \textbf{81.6} & 68.7 & 74.0 & 70.3 & 87.6 & 68.5 & \textbf{88.9} & 80.0    \\ 
		PointNet++ \cite{qi2017pointnet++}  & 79.5 & 71.6 & \textbf{87.7} & \textbf{70.7} & \textbf{88.8} & 64.9 & 88.8 & 78.1  \\
		PointCNN \cite{li2018pointcnn}  & 78.0 &80.1 &78.2 &68.2 & 81.2 & 70.2 &82.0 &70.6 \\
		DGCNN \cite{wang2018edgeconv}  & 77.7 & 71.8 & 77.7 & 55.2 & 87.3 & 68.7 & 88.7 & 85.5 \\  
		SpiderCNN \cite{xu2018spidercnn}  & 74.3 & 72.4 & 72.6 & 58.4 & 82.0 & 68.5 & 87.8 & 81.3 \\
		RS-CNN \cite{liu2019relation}  & 71.8 & 76.4 & 78.9 & 68.1 & 80.2 & 62.5 & 82.6 & 76.6 \\
		RIConv ~\cite{zhang-riconv-3dv19}  & 80.6 & 80.2 & 70.7 & 68.8 & 86.8 & 70.4 & 87.2 & 84.3 \\
		Ours  & 81.2 & \textbf{82.6} & 81.6 & 70.2 & 88.6 & \textbf{70.6} & 86.2 & \textbf{86.6} \\
		\midrule
		Network   & lamp & laptop & motor & mug & pistol & rocket & skate & table \\ 
		\midrule 
		PointNet \cite{qi2017pointnet}   & 74.9 & 83.6 & 56.5 & 77.6 & 75.2 & \textbf{53.9} & \textbf{69.4} & 79.9   \\ 
		PointNet++ \cite{qi2017pointnet++} & 79.2 & \textbf{94.9} & 54.3 & \textbf{92.0} & 76.4 & 50.3 & 68.4 & 81.0   \\
		PointCNN \cite{li2018pointcnn} & 68.9 & 80.8 &48.6 &77.3 &63.2 &50.6 &63.2 & \textbf{82.0} \\
		DGCNN \cite{wang2018edgeconv} &\textbf{81.8} & 81.3 & 36.2 & 86.0 & 77.3 & 51.6 & 65.3 & 80.2 \\  
		SpiderCNN \cite{xu2018spidercnn} & 71.3 & 94.5 & 45.7 & 88.1 & \textbf{83.4} & 50.5 & 60.8 & 78.3  \\
		RS-CNN \cite{liu2019relation} & 73.2 & 90.2 & 54.8 & 89.8 & 72.8 & 43.6 & 65.3 & 72.6  \\
		RIConv ~\cite{zhang-riconv-3dv19} & 78.0 & 80.1 & 57.3 & 91.2 & 71.3 & 52.1 & 66.6 & 78.5 \\
		Ours & 81.6 & 79.6 & \textbf{58.9} & 90.8 & 76.8 & 53.2 & 67.2 & 81.6 \\
		\bottomrule
	\end{tabular}
	\caption{Per-class accuracy of object part segmentation on the ShapeNet dataset in SO3/SO3 scenario. Our method works equally well to previous methods in this scenario.}
	\label{tab:objectpart_perclass_so3_so3}
\end{table*}

\begin{table*}[t]
	\centering
	\begin{tabular}{p{80pt} p{30pt}p{30pt}p{30pt}p{30pt}p{30pt}p{30pt}p{30pt}p{30pt}p{30pt}p{30pt}p{30pt}p{30pt}p{30pt}p{30pt}p{30pt}p{30pt}}
		\toprule
		Network   & aero & bag & cap & car & chair & earph. & guitar & knife \\ 
		\midrule 
		PointNet \cite{qi2017pointnet}  & 40.4 & 48.1 & 46.3 & 24.5 & 45.1 & 39.4 & 29.2 & 42.6   \\ 
		PointNet++ \cite{qi2017pointnet++}  & 51.3 & 66.0 & 50.8 & 25.2 & 66.7 & 27.7 & 29.7 & 65.6 \\
		PointCNN \cite{li2018pointcnn}  &21.8 & 52.0 &52.1 &23.6 &29.4  &18.2 &40.7 &36.9 \\
		DGCNN \cite{wang2018edgeconv} & 37.0 & 50.2 & 38.5 & 24.1 & 43.9 & 32.3 & 23.7 &48.6 \\ 
		SpiderCNN \cite{xu2018spidercnn}  & 48.8 & 47.9 & 41.0 & 25.1 & 59.8 & 23.0 & 28.5 & 49.5  \\
		RS-CNN \cite{liu2019relation}  & 26.9 & 49.7 & 44.7 & 25.3 & 36.5 & 30.0 & 33.3 & 39.4  \\
		RIConv ~\cite{zhang-riconv-3dv19}  & 80.6 & 80.0 & 70.8 & 68.8 & 86.8 & 70.3 & 87.3 & 84.7 \\
		Ours  & \textbf{80.9} & \textbf{82.6} & \textbf{81.0} & \textbf{70.2} & \textbf{88.4} & \textbf{70.6} & \textbf{87.1} & \textbf{87.2} \\
		\midrule 
		Network   & lamp & laptop & motor & mug & pistol & rocket & skate & table \\ 
		\midrule 
		PointNet \cite{qi2017pointnet} & 52.7 & 36.7 & 21.2 & 55.0 & 29.7 & 26.6 & 32.1 & 35.8   \\ 
		PointNet++ \cite{qi2017pointnet++}  & 59.7 & 70.1 & 17.2 & 67.3 & 49.9 & 23.4 & 43.8 & 57.6   \\
		PointCNN \cite{li2018pointcnn}  &51.1 &33.1 &18.9 &48.0 &23.0 &27.7 &38.6 &39.9 \\
		DGCNN \cite{wang2018edgeconv} &54.8 & 28.7 & 17.8 & 74.4 & 25.2 & 24.1 & 43.1 & 32.3 \\ 
		SpiderCNN \cite{xu2018spidercnn}  & 45.0 & \textbf{83.6} & 20.9 & 55.1 & 41.7 & 36.5 & 39.2 & 41.2  \\
		RS-CNN \cite{liu2019relation}  & 54.9 & 36.1 & 20.6 & 53.3 & 29.0 & 29.4 & 32.3 & 42.6  \\
		RIConv ~\cite{zhang-riconv-3dv19}  & 77.8 & 80.6 & 57.4 & \textbf{91.2} & 71.5 & 52.3 & 66.5 & 78.4 \\
		Ours & \textbf{81.8} & 78.9 & \textbf{58.7} & 91.0 & \textbf{77.9} & \textbf{52.3} & \textbf{66.8} & \textbf{80.3} \\
		\bottomrule
	\end{tabular}
	\caption{Per-class accuracy of object part segmentation on the ShapeNet dataset in z/SO3 scenario. Our method significantly outperforms previous methods thanks to the rotation invariance features from our convolution operators.}
	\label{objectpart_perclass_z_so3}
\end{table*}

\end{document}